\documentclass{article} 
\usepackage{iclr2027_conference,times}

\usepackage{amsmath,amsfonts,bm}

\def\eqref#1{equation~\ref{#1}}

\def\1{\bm{1}}

\DeclareMathAlphabet{\mathsfit}{\encodingdefault}{\sfdefault}{m}{sl}
\SetMathAlphabet{\mathsfit}{bold}{\encodingdefault}{\sfdefault}{bx}{n}

\usepackage{hyperref}
\usepackage{url}
\usepackage{booktabs}
\usepackage{graphicx}
\usepackage[T1]{fontenc}
\graphicspath{{figures/}}

\usepackage{xcolor}
\usepackage[frozencache,cachedir=.]{minted}

\setminted{escapeinside=||}

\newcommand{\codemethod}[1]{\textcolor{pyString}{#1}}

\definecolor{pyKeyword}{RGB}{0,92,197}
\definecolor{pyComment}{RGB}{110,110,110}
\definecolor{pyString}{RGB}{0,128,128}
\definecolor{pyBuiltin}{RGB}{170,0,170}
\definecolor{pyProject}{RGB}{180,90,0}
\definecolor{codeBg}{RGB}{247,247,247} 
\definecolor{codeRule}{RGB}{210,210,210} 
\setminted{fontsize=\small,breaklines=true,linenos=true,frame=leftline,framesep=3pt,rulecolor=codeRule,fontfamily=tt,fontseries=b}

\title{Jaxnasium: Embracing PyTrees For Any-Agent, Cross-Suite Reinforcement Learning}

\author{Koen Ponse, Aske Plaat \& Thomas Moerland \\
Leiden Institute of Advanced Computer Science \\
Leiden University\\
\texttt{\{k.ponse,a.plaat,t.m.moerland\}@liacs.leidenuniv.nl} 
}

\definecolor{noteBlue}{RGB}{0,102,255}

\iclrfinalcopy 
\begin{document}

\maketitle
\lhead{}

\begin{abstract}

    Hardware-accelerated reinforcement learning in JAX has gained rapid popularity in recent years and has proven to be a successful method of scaling online reinforcement learning.
    The resulting ecosystem, however, is fragmented: environment suites
define their own interfaces, and algorithm implementations are typically written
for a single suite, a single type of space, and either JAX-native or host-based
environments.
    Here we present Jaxnasium, a library for JAX-based reinforcement learning, aiming to make the field more accessible.
    Jaxnasium represents observation and action spaces as PyTrees, and uses this structure to build networks and support multi-agent processing.
    Through this, its near-single-file implementations run on discrete, continuous, image, and composite spaces, as well as in multi-agent environments, without any changes to the algorithm code.
    The same files serve as readable, research-friendly algorithms, as well as importable algorithms for more applied workloads. Through a minimal Gymnasium-style interface and wrappers for 11 suites, over 300 environments are available on which a small PPO implementation with default hyperparameters is able to run. Additionally, through a simple wrapper, we can run Gymnasium and EnvPool environments with the same algorithm and training loop, with no performance loss compared to architectures specifically built for training JAX algorithms on host-based environments. Jaxnasium further provides tools for multi-seed evaluation, hyperparameter sweeps, and project scaffolding. Jaxnasium is open-sourced at \url{https://github.com/ponseko/jaxnasium}.

\end{abstract}

\section{Introduction}

Running the entire reinforcement learning loop on the accelerator has reshaped what is possible for many researchers. When both the environment and the agent are written in JAX~\citep{jax2018github}, rollouts and updates can together be compiled and vectorized without host (CPU) round-trips, which can yield speed-ups of several orders of magnitude~\citep{hessel2021podracer,lu2022discoveredpurejaxrl}. A broad ecosystem of
JAX-native environments has emerged, including Gymnax~\citep{gymnax2022github},
Brax~\citep{brax2021github}, Jumanji~\citep{bonnet2024jumanji},
Pgx~\citep{koyamada2023pgx}, JaxMARL~\citep{flair2024jaxmarl},
XLand-MiniGrid~\citep{nikulin2023xlandminigrid}, and Craftax~\citep{matthews2024craftax}, each with its own interface and conventions for state, time steps and spaces. This expansion added to the already large landscape of reinforcement learning.

Algorithm implementations, however, have not kept pace with this breadth. They
tend to be written for a specific environment suite and/or space type. 
As a result, even within the same library, an algorithm is often re-implemented for discrete and continuous actions, for image and vector observations, for single- and multi-agent settings. 
Libraries serve different users in different ways: research-oriented libraries provide readable single- or near-single-file implementations that are easy to
modify~\citep{huang2022cleanrl,lu2022discoveredpurejaxrl},
while production-oriented libraries~\citep{stable-baselines3,liang2018rllib} offer
a single importable algorithm that handles many environments, at the cost of
spreading the implementation over many interacting and inherited components, making them less accessible for researchers who want to adapt the underlying algorithms. 

A new divide now separates JAX-native environments from the many environments that still run on the host CPU, such as those in Gymnasium~\citep{neurips_gymnasium} and
EnvPool~\citep{weng2022envpool}. For the former, the entire training loop can run on
the accelerator; for the latter, the environment steps on the host and only
the network runs on the accelerator. These two settings call for different
training architectures (originally proposed as Anakin and Sebulba in
\citet{hessel2021podracer}), and different implementations are then again written for each of them. 
As a result, the starting point for researchers is typically an implementation that supports a subset of tasks or is difficult to modify.

We introduce \textbf{Jaxnasium}, a lightweight RL library built on
Equinox~\citep{kidger2021equinox}, with the aim of making JAX-based RL more accessible. At its core are near-single-file algorithm
implementations that operates many environments. They build on
PyTrees, JAX's notion of arbitrarily nested structures (including structures built with standard Python containers such as dictionaries, tuples and lists). In Jaxnasium, observation and action spaces are PyTrees of primitive spaces, and each algorithm constructs its networks from the structure of these spaces without any complicated machinery. A single
implementation trains on discrete, continuous, image and composite
spaces alike, and even seamlessly extends to multi-agent settings, where each agent simply has its own entry in the space PyTree. 
Because JAX provides simple operations on PyTrees, this generally requires little special-case code, and our implementations remain readable and modifiable for algorithm researchers. 

\begin{figure}[h]
\centering
\begin{minipage}[c]{0.60\linewidth}
\begin{minted}[xleftmargin=1.2em,numbersep=6pt]{python}
import jaxnasium as jym
import jaxnasium.algorithms as jxa

ppo = jxa.|\codemethod{PPO}|()
envs = jym.|\codemethod{registry}|.registered_envs

for env_id in envs:
    env = jym.|\codemethod{make}|(env_id)
    agent, metrics = ppo.|\codemethod{train}|(key, env)

    # agent.save(...) 
    # agent.evaluate(...)

\end{minted}
\end{minipage}\hfill%
\begin{minipage}[c]{0.38\linewidth}
\centering
\includegraphics[width=\linewidth]{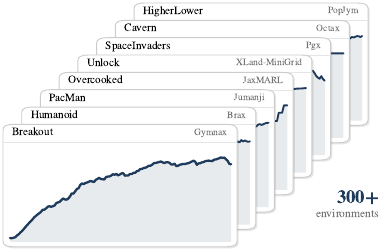}
\end{minipage}
\caption{Jaxnasium registers 333 environments and includes algorithm implementations that can train on over 300 of them without any changes. The functionality is familiar to those who have used well-known libraries such as Gymnasium and Stable-Baselines.}
\label{fig:jaxnasium-overview}
\end{figure}

Building on Equinox further lets us write
algorithms as classes while respecting JAX's functional requirements, giving
practitioners the familiar interface of libraries such as
Stable-Baselines3. To let these algorithms run on many environments, Jaxnasium
defines a minimal, Gymnasium-style environment interface and currently wraps over
300 environments from different domains into it, on which a single small PPO implementation can operate
(Figure~\ref{fig:jaxnasium-overview}). Although the algorithms target end-to-end JAX training
for maximum performance, host-based environments (like Gymnasium and
EnvPool) run within the same compiled training loop through a wrapper rather than
a separate implementation, matching or exceeding the throughput of
implementations built specifically for host-based environments
(Section~\ref{sect:environment-interop}).

Concretely, our contributions are:
\begin{itemize}
  \item \textbf{Space-agnostic algorithms.} Near-single-file implementations whose networks are constructed from the space
    PyTrees, so that each algorithm supports arbitrary observation and action
    spaces and extends to multi-agent settings without code changes. The same
    code serves as a readable reference for researchers and as an importable,
    Stable-Baselines-style algorithm for practitioners.
  \item \textbf{One implementation for JAX-native and host environments.}
    Gymnasium and EnvPool environments run inside the same compiled training
    loop as JAX-native ones. We show that this loses no throughput compared to
    implementations dedicated to either setting.
  \item \textbf{A common environment interface.} A minimal, Gymnasium-style
    functional interface with PyTree spaces, and wrappers that bring 11
    existing JAX environment suites under it.
  \item \textbf{Research tooling.} A command-line interface that scaffolds new
    projects, and utilities for multi-seed evaluation and hyperparameter sweeps
    (grid, random, Sobol and one-at-a-time), and a small \texttt{tree} package for convenience operations over PyTrees.
\end{itemize}

\section{Background \& Landscape}

Reinforcement learning has notoriously been data-hungry~\citep{kaiser2024modelbased} and as a result, accelerating data generation has been an important challenge in the field. Because early environment suites had been implemented primarily in Python, a natural optimization strategy has been to (re)implement environments in faster compiled languages such as C or C++~\citep{weng2022envpool, cobbe2019procgen, suarez2025pufferlib, shacklett23madrona}. 

While these efforts dramatically reduce the CPU-side overhead, they cannot address the fundamental issue that data has to repeatedly travel between the host CPU (where trajectories are being generated in the environment), and the accelerator device (where the policy generates actions and the networks are updated). Additionally, as accelerators are only part of the reinforcement learning process, the field essentially ``lost the hardware lottery''~\citep{MLST}, as it struggled to benefit from the massive GPU scaling that propelled computer vision and natural language processing following the ImageNet moment~\citep{5206848_imagenet, NIPS2012_alexnet}.

To bypass the communication bottleneck, one may attempt to execute the environment transitions directly on the accelerator. WarpDrive~\citep{JMLR:v23:22-0185} demonstrated this capability by writing the environment dynamics directly in CUDA kernels. However, writing custom CUDA kernels introduced engineering friction for the machine learning community. JAX~\citep{jax2018github} emerged as a compelling alternative, allowing researchers to write familiar high-level Python code that could automatically be vectorized and compiled to accelerators via XLA.

\paragraph{End-to-End Reinforcement Learning in JAX} 
Through JAX, accelerator-native environments enabled unified training pipelines without device communication overhead.~\citet{hessel2021podracer} formalized this idea for reinforcement learning and introduced the \textbf{Anakin} architecture, which trained agents end-to-end on accelerators when environments were JIT-compatible through JAX. The same paper also introduced the \textbf{Sebulba} architecture for traditional environments in which data generation is still executed on the host device. Such an architecture was (and is) still relevant, as the vast majority of environments are not written to be fully JIT-compatible.

The Anakin architecture was later popularized as PureJAXRL~\citep{lu2022discoveredpurejaxrl}, which provided simple single-device implementations (in the spirit of CleanRL~\citep{huang2022cleanrl}) for algorithms such as PPO~\citep{schulman2017proximalpolicyoptimizationalgorithms}. They showed how this method was able to train reinforcement learning agents multiple orders of magnitude faster, possibly reducing experiment training times from days to minutes.

\paragraph{Functional Environments}
Standard reinforcement learning APIs, most prominently Gymnasium~\citep{neurips_gymnasium} and \texttt{dm-env}~\citep{dm_env2019}, rely on stateful object-oriented designs with in-place mutations. Because JAX relies on pure, side-effect-free functions, the traditional interface was incompatible with PureJAXRL-style training. Although Gymnasium eventually introduced an experimental functional interface \texttt{FuncEnv}, it arrived after several JAX-native suites~\citep{gymnax2022github, brax2021github} had been available, and its supporting utilities are not designed to be JIT-compatible. As a result, this interface has seen little adoption.

\paragraph{JAX}

JAX~\citep{jax2018github} is a Python library focused on efficient array operations and built around composable function transformations. These function transformations are most notably Just-In-Time compilation via XLA (\texttt{jax.jit}), automatic vectorization (\texttt{jax.vmap}), and automatic differentiation (\texttt{jax.grad}). In reinforcement learning, \texttt{jax.jit} allows an entire training loop to be
compiled for the accelerator, and \texttt{jax.vmap} turns a single-environment
step function into a batched one without modification.
These function transformations, however, come with constraints: the function inputs and outputs must be numerical arrays or scalars, and those arrays must have shapes that are known at compile time; and finally, the functions must be pure, free of side-effects. Any state, like the environment state, must therefore be passed into a function explicitly and returned from it.

\paragraph{PyTrees}
Such state is rarely just a single array. To accommodate more complex structures of arrays, JAX introduces the concept of PyTrees and allows PyTrees of numerical arrays as valid inputs for its transformations. A PyTree is a (nested) tree structure whose internal nodes are registered containers (such as \texttt{tuple} and \texttt{dict} by default) and whose leaves are anything not registered as such a container.
JAX operates over these PyTrees through defined flattening and unflattening rules, which decompose a tree into a flat sequence of leaves (on which to operate) alongside the structure metadata required for reconstruction.
Outside of transformations, PyTree utilities such as
\texttt{jax.tree.map} accept leaves of any type, which provides a convenient way to operate over any nested structure (Section~\ref{subsec:pytrees-of-spaces}).

\paragraph{Equinox} To bridge between JAX's functional paradigm and object-oriented code, \citet{kidger2021equinox} introduced Equinox. Its core building block, \texttt{equinox.Module}, is an enhanced Python dataclass that is automatically registered as a JAX PyTree. Within a module, array attributes (such as neural network weights) are treated as transformable PyTree leaves, while non-array attributes (such as configuration metadata) can be marked as part of the static tree structure or filtered out. Because Equinox modules are valid JAX PyTrees, methods defined on them remain functional, and instances can freely cross \texttt{jax.jit}, \texttt{jax.vmap}, and \texttt{jax.grad} boundaries. Jaxnasium adopts this object-functional architecture and builds its environments,
wrappers, networks and algorithms as Equinox modules.

\section{Environments as PyTrees} \label{sect:env}

Currently, many JAX environment suites have been created, and many have established their own JAX-compatible interface. Although some suites borrowed their interface from existing libraries~\citep{matthews2024craftax, tomilin2026crax, morad2023popgym}, many suites have opted to create their own~\citep{gymnax2022github, bonnet2024jumanji, brax2021github, koyamada2023pgx, flair2024jaxmarl, nikulin2023xlandminigrid, pignatelli2024navix, radji2025octax}. Similar to traditional suites, multi-agent reinforcement learning has also received its own functional interface~\citep{flair2024jaxmarl, dekock2023mava}, primarily following the PettingZoo~\citep{terry2021pettingzoo} convention. Using several suites
from the same code therefore requires wrappers that bridge these
differences~\citep{toledo2025stoa}.

Jaxnasium is not an environment suite. The interface introduced here primarily exists to provide the algorithms of Section~\ref{sect:algorithms} a single target. Those algorithms primarily require: i) a functional, JIT-compatible signature; ii) separate
termination and truncation signals, so that value estimates can be bootstrapped
correctly; and iii) observation and action spaces that are arbitrary PyTrees, so that
composite spaces and multi-agent settings share one abstraction. Beyond these
requirements, we made two choices for convenience: the interface stays close to
Gymnasium~\citep{neurips_gymnasium}, so that it is familiar to most practitioners, and environments are
PyTrees themselves, so that their parameters travel with the object and they can freely pass through JAX function transformations. 
Section~\ref{sect:environment-interop} describes how existing suites are wrapped
into this interface.

\subsection{Environment Interface} \label{subsec:interface}

Jaxnasium largely follows the functional design of Gymnax~\citep{gymnax2022github}: \texttt{reset} returns an initial
observation together with the environment state, and \texttt{step} takes that
state explicitly and returns the next one. For the output of \texttt{step}, we primarily mirror Gymnasium~\citep{neurips_gymnasium} and separate natural termination from truncation, e.g.\ by a time limit. \footnote{Recent versions of Gymnax (from v1.0) likewise return separate termination and truncation flags.}
\begin{equation*}
    (\text{obs}, \text{reward}, \text{terminated}, \text{truncated}, \text{info}), s_{t+1} \leftarrow \texttt{step}(\text{key}, s_t, a)
\end{equation*}
Environments reset automatically when an episode ends, as is usual in
vectorized training. Since the returned observation then belongs to
the next episode, the final observation of the finished episode is stored in
\texttt{info}, so that algorithms can still correctly bootstrap from it after truncation. Listing~\ref{lst:env-usage} shows the resulting
interaction loop.

Gymnax passes the environment parameters as a separate argument to \texttt{step} and
\texttt{reset} and treats the environment object itself as static under
\texttt{jax.jit}. Jaxnasium instead defines environments and wrappers as Equinox
modules~\citep{kidger2021equinox}, such that parameters are attributes on the object.
An environment is then itself a PyTree and can be passed through \texttt{jax.jit}
and \texttt{jax.vmap} like any other value. For settings that require changing parameters on every call, such as meta-RL, we provide a wrapper for the Gymnax-style parameter argument.

\begin{listing}[t]
\begin{minted}[xleftmargin=1em,numbersep=6pt]{python}
import jax
import jaxnasium as jym

key = jax.random.PRNGKey(0)
env = jym.|\codemethod{make}|("CartPole-v1")

key, reset_key, step_key = jax.random.|\codemethod{split}|(key, 3)
obs, state = env.|\codemethod{reset}|(reset_key)
action = env.action_space.|\codemethod{sample}|(step_key)

(obs, reward, terminated, truncated, info), state = env.|\codemethod{step}|(
    step_key, state, action
)
\end{minted}
\caption{Interacting with a Jaxnasium environment. Apart from the explicit PRNG
keys and environment state required by JAX, the calls mirror Gymnasium.}
\label{lst:env-usage}
\end{listing}

\subsection{PyTrees of Spaces} \label{subsec:pytrees-of-spaces}

For environment space specifications, Jaxnasium provides standard observation and action spaces in the form of \texttt{Box}, \texttt{Discrete}, and \texttt{MultiDiscrete}, preserving the same abstractions that have long been established and understood~\citep{1606.01540}. Jaxnasium only deviates from Gym(nasium) spaces for some methods, e.g.\ to support JIT-compatible sampling.

The bigger architectural change in Jaxnasium occurs in composite spaces. In Gymnasium, composite spaces are defined via dedicated container classes such as \texttt{gymnasium.spaces.Dict} or \texttt{gymnasium.spaces.Tuple}, which requires algorithms to understand and unpack these types of spaces within their training pipelines. Instead, Jaxnasium has no composite space classes and instead embraces JAX's PyTree abstraction: \emph{any arbitrary PyTree whose leaves are primitive spaces is a valid composite space}. Because Python dictionaries, tuples, and lists are already registered PyTrees, users can structure complex observation and action spaces using standard Python idioms, and operations on a composite space are then simply a mapping over the tree:

\begin{minted}[xleftmargin=1em,numbersep=6pt]{python}
import jax
import jaxnasium as jym

action_space = {
    "arrow_keys": jym.spaces.Discrete(4),
    "mouse": (
        jym.spaces.Box(-1.0, 1.0, (2,)), 
        jym.spaces.Discrete(2)
    ),
}
keys = jym.tree.split_key_like(key, action_space)
jax.tree.map(lambda s, k: s.sample(k), action_space, keys)
# > {"arrow_keys": 2, "mouse": (Array([0.2, -0.4]), 1)}
\end{minted}

This choice has two advantages for algorithms: 1) because the observation and action spaces are PyTrees, networks can, without any special-case code, construct matching structured networks; observation leaves can produce encoders, and action leaves can produce output heads (Section~\ref{sect:algorithms-nets}). 2) we need no separate abstraction or interface for multi-agent environments. A multi-agent environment is simply an environment where spaces and returned rewards are PyTrees with one entry per agent at the top level of the PyTree. Agents may have different spaces, and these spaces may themselves be composite. Section~\ref{sect:algorithms-ma} shows how algorithms use this structure.

\section{Algorithms over PyTrees} \label{sect:algorithms}

As outlined in the introduction, reinforcement learning libraries tend to offer either research-focused, readable implementations of single- or near-single-file algorithms~\citep{huang2022cleanrl,
lu2022discoveredpurejaxrl}, or production-focused importable algorithms that handle more environment types from the same exported class~\citep{stable-baselines3, rl-zoo3, liang2018rllib}.
Jaxnasium aims to offer both from the same code. We provide small implementations of PPO~\citep{schulman2017proximalpolicyoptimizationalgorithms}, SAC~\citep{haarnoja2018sac}, DQN~\citep{mnih2013dqn}, and PQN~\citep{gallici2025pqn}, separating only networks and container objects such as replay buffers and normalization classes. 

In addition, however, while retaining a fully JIT-compatible training loop for end-to-end reinforcement learning~\citep{lu2022discoveredpurejaxrl, hessel2021podracer}, Jaxnasium algorithms are built as importable classes (Equinox Modules~\citep{kidger2021equinox}) with the library being available through pip. As such, similar to the second category of libraries, Jaxnasium algorithms are usable through a friendly API similar to Stable-Baselines. Algorithms are implemented such that they can learn over various action spaces, and because their networks are separated and constructed over the PyTree of spaces (Sections~\ref{subsec:pytrees-of-spaces} and~\ref{sect:algorithms-nets}), all these algorithms can operate over almost any type of environment.

\subsection{PyTrees of Networks} \label{sect:algorithms-nets}

\begin{figure}[h]
    \centering
    \includegraphics[width=0.85\linewidth]{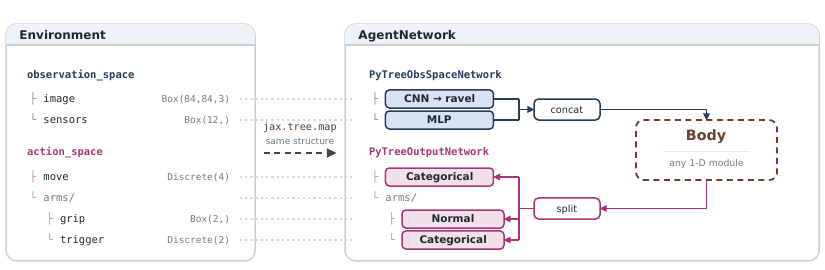}
    \caption{The default agent (policy) network. An observation network creates an
    encoder for each observation leaf, a shared body processes the concatenated
    features, and an output network creates a head for each action leaf, with a
    distribution matching the leaf's space. While the structure is useful for mapping spaces onto networks, it remains optional. Critic networks are created similarly, but with different output heads.}
    \label{fig:agent-network}
\end{figure}

Because Jaxnasium represents composite spaces as native PyTrees (Section~\ref{subsec:pytrees-of-spaces}), neural network architectures can mirror this tree structure by simply mapping over the tree structure. The default agent network of Jaxnasium consists of three parts
(Figure~\ref{fig:agent-network}). An observation network maps over the
observation space and creates an encoder for each leaf. The resulting features are concatenated and passed
through a shared body, by default a SimBa~\citep{lee2025simba} network. An output
network then maps over the action space and creates a head for each leaf, with a
distribution matching its type, e.g.\ categorical for discrete and Gaussian for
continuous actions.

While this is how algorithms initialize themselves by default, each part of the network can be replaced and we provide additional architectures to insert into the agent networks, such as BroNet~\citep{nauman2024bronet}, a CNN, or an MLP. However, the algorithms themselves only expect a callable module and as such, the default agent network is purely optional. For those more invested in the Flax~\citep{flax2020github} ecosystem, we provide examples in our documentation on how to use custom Flax networks within a Jaxnasium algorithm.

\subsection{PyTrees of Agents (Multi-Agent)} \label{sect:algorithms-ma}

This PyTree formulation for spaces extends naturally to multi-agent reinforcement learning, without any new abstractions. As described in
Section~\ref{subsec:pytrees-of-spaces}, a multi-agent environment exposes its
observations, actions and rewards as PyTrees whose first level indexes the agents. When an algorithm is trained on such an environment, it creates one agent per entry of this first level, each seeing a single-agent view of the environment with only its own spaces. Typical operations on a single agent are now automatically dispatched over each agent; the transformation to multi-agent execution occurs purely at the PyTree boundary. As a result, the entire algorithm and agent logic can remain written as single-agent code, coupling these two fields back together into the same programming model.

When all agents have identical shapes, their operations are vectorized with
\texttt{jax.vmap}; otherwise, they run per agent. Hyperparameters can be set per
agent by passing a PyTree with the agent structure, as long as they do not change
the structure of the shared training loop (Listing~\ref{lst:multi-agent}).

\begin{listing}[t]
\begin{minted}[xleftmargin=1em,numbersep=6pt]{python}
env = jym.make("MPE_simple_spread_v3")
env.action_space
# > {"agent_0": Discrete(5), "agent_1": Discrete(5), ...}

ppo = jxa.PPO(gamma={"agent_0":0.99, "agent_1":0.99, "agent_2":0.95})
agent, metrics = ppo.train(key, env)  # one PPO agent per entry
agent.get_action(key, obs)  
# > {"agent_0": ..., "agent_1": ..., ...}
\end{minted}
\caption{Training PPO on a multi-agent environment. The algorithm is used
unchanged; the per-agent discount factors are optional and illustrate
heterogeneous configuration.}
\label{lst:multi-agent}
\end{listing}

By default, our multi-agent transformation yields independent learners: each agent
optimizes its own objective with its own networks (IPPO~\citep{dewitt2020ippo}).
Centralized critic observations in the style of MAPPO are additionally supported, and we provide documentation examples on mirroring networks across agents.

\section{Environment Interoperability} \label{sect:environment-interop}

Similar to earlier efforts~\citep{toledo2025stoa}, to allow the algorithms of Section~\ref{sect:algorithms} to train seamlessly on as many environments as possible, Jaxnasium provides wrappers to convert many existing environment suites to the expected interface (Section~\ref{subsec:interface}). In addition to wrappers for existing JAX suites, which allow training pipelines to be fully compiled to the accelerator, Jaxnasium provides wrappers for host-based environments, which run Gymnasium~\citep{neurips_gymnasium} and EnvPool~\citep{weng2022envpool} environments from within the same compiled training loop. As such, rather than requiring separate code implementations for host-bound and JAX-native setups, Jaxnasium consolidates the two regimes in a single simple research-friendly implementation.

\paragraph{JAX-Native Suites} \label{sect:interop-jax}

Jaxnasium currently registers 333 environments from eleven existing JAX suites, and exposes a familiar \texttt{make} command to initialize each environment and wrap it automatically with the appropriate wrapper. The wrappers translate each suite's signature, time steps and spaces into the Jaxnasium interface, and pass on
action masks where a suite provides them. A full list of current environment suites, covering various domains, is given in Appendix~\ref{app:environment-list}. Included algorithms can operate on over 300 of these environments directly in an end-to-end training loop (Figure~\ref{fig:jaxnasium-overview}). However, the environment registry and wrappers may also be consumed without Jaxnasium algorithms. While more wrappers are likely to be included in the future, new environments may also be created directly as Jaxnasium environments \citep{ponse2025chargax,ponseeconojax,zhang2025ai}.

\paragraph{Host-Based Environments} \label{sect:interop-host}

Many reinforcement learning environments are not written in JAX and excluding them would omit a large number of possible benchmarks. Rather than providing a separate implementation to run these host-based environments~\citep{hessel2021podracer, huang2022cleanrl, toledo2024stoix}, Jaxnasium provides a wrapper to run (vectorized) Gymnasium~\citep{neurips_gymnasium} or EnvPool environments~\citep{weng2022envpool} to transition these environments through JAX callbacks. The wrapper converts the host spaces to
Jaxnasium spaces and extracts the final observation of finished episodes, so
that automatic resets and bootstrapping behave as they do for JAX-native
environments. EnvPool environments can alternatively be stepped through
EnvPool's own XLA interface. To the algorithm, the wrapped environment is an ordinary environment: rollout,
update and network code are unchanged, and only the environment transition runs
on the host. 

\paragraph{Performance}
Although end-to-end training in JAX using native JAX environments is known to boost performance, it is reasonable to ask whether our simplified callback wrapper approach sacrifices performance compared with keeping the implementations separate.
To test this, we compare PPO throughput against Stable-Baselines3~\citep{stable-baselines3},
the PyTorch and JAX implementations of CleanRL~\citep{huang2022cleanrl}, and the
Anakin and Sebulba implementations of Stoix~\citep{toledo2024stoix}, on four
backends: JAX-native environments, Gymnasium, EnvPool, and EnvPool through its
XLA interface. All implementations use the same hyperparameters: rollouts of
4096 transitions, 4 epochs of 4 minibatches, $64\times64$ networks, and
observation and reward normalization. Each
configuration trains for $2\cdot10^5$ environment steps with 8, 512 or 2048
parallel environments. We report environment steps per second over the full
training run, excluding compilation, as the mean of five runs after one
discarded warm-up run, and take the geometric mean over the five
classic-control environments and Catch. All runs use a single NVIDIA RTX 5090.

\begin{figure}[h]
  \centering
  \includegraphics[width=\linewidth]{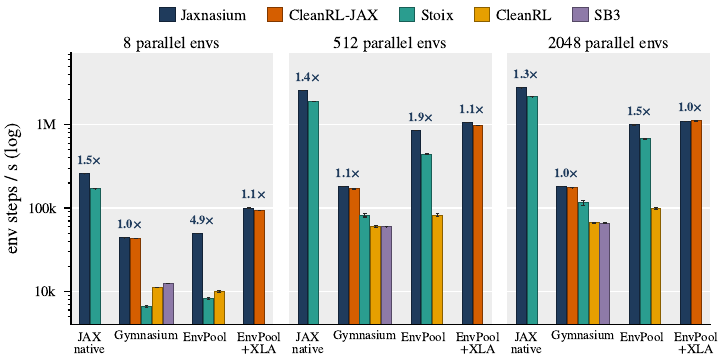}
  \caption{Average throughput of Jaxnasium and other existing established libraries on various environment backends with a small PPO network for each. Not all libraries have an implementation available to run on each environment backend, hence these bars are omitted from the graph. Jaxnasium, despite using a single small implementation to run on all types of environments, matches or exceeds the throughput of other libraries. Because not many environments have been implemented in all these backends, this graph only averages over 6 simple environments, with a cheap step function. Appendix~\ref{app:performance-extra} lists extra environments, which paint roughly the same picture.}
  \label{fig:benchmark-gymnasiumwrapper}
\end{figure}

The results of these runs are reported in Figure~\ref{fig:benchmark-gymnasiumwrapper}. Jaxnasium matches or exceeds the fastest baseline in every environment backend. For the settings we tested, a separate implementation for host-based environments is therefore not necessary. We did not test multi-device setups, where asynchronous designs such as Sebulba~\citep{toledo2024stoix} may pay off, but believe this shows Jaxnasium can be a good starting point for the compute available to many research labs. The absolute throughput per library on further (more complex) environments is given in Appendix~\ref{app:performance-extra}.

\section{Research Tooling}

Beyond environments and algorithms, running reinforcement learning experiments
involves recurring work: setting up a project, evaluating over many seeds, and
sweeping hyperparameters across environments and algorithms. Jaxnasium includes
lightweight tools for each of these.
While many of these recurring tasks, particularly project setup and setting up evaluations, are increasingly generated by large language models, details
such as seeding and aggregation over runs can affect experimental conclusions. We hope that providing default implementations for these tasks provides researchers with tools to create a more controlled and deterministic result.

\paragraph{Project Scaffolding with \texttt{create-rl-app}}

Inspired by the web development community, Jaxnasium ships a command-line interface, \texttt{create-rl-app}, that scaffolds new projects and copies templates into existing ones. This offers a simplified way to use Jaxnasium environment templates or algorithms as a starting point for further research.

\begin{minted}[xleftmargin=1em,numbersep=6pt]{bash}
uvx jaxnasium init          # new project with training script
uvx jaxnasium add ppo       # ppo.py and its networks, ready to modify
uvx jaxnasium add sweep     # template for a multi-environment sweep
\end{minted}

While \texttt{create-rl-app} is integrated into Jaxnasium, it is released as a standalone package and may be included by anyone to create a CLI to share their templates, simply by defining templates in a configuration \texttt{toml} file.

\paragraph{Evaluation and Sweeps} \label{sect:eval}

Reliable conclusions in reinforcement learning require many seeds and careful
aggregation~\citep{henderson2019deepreinforcementlearningmatters, agarwal2021deep}. Jaxnasium's evaluation module provides intuitive tooling for sweeping over arguments of any function; typically a training run over various algorithms, hyperparameters, and/or environments. The module provides several standard search methods such as grid search, random search, and Sobol search, as well as evaluation metrics for reinforcement learning~\citep{agarwal2021deep}. An example sweep configuration is given in Appendix~\ref{app:sweep}.

\paragraph{Other Utilities}
The \texttt{jaxnasium.tree} module offers operations over PyTrees that JAX does
not provide directly but that are common in our reinforcement learning workload, such as stacking and unstacking trees, mapping over the
first level of a tree (the agent level in multi-agent settings), and splitting
PRNG keys to match a tree's structure. Furthermore, environment wrappers cover vectorization,
episode logging, observation and reward normalization, observation flattening
and discretization of continuous action spaces. Trained agents can be saved and loaded and have normalization built in, and training functions can be compiled ahead of time to separate compilation from run time.

\section{Limitations \& Future Work}

Currently, Jaxnasium implements four algorithms. While we aim to include more algorithms in the future, we also aim to update algorithms with newly found improvements, rather than sticking to the exact original implementation. As a result, to keep the library maintainable, Jaxnasium is unlikely to offer as many algorithms as some other libraries do~\citep{toledo2024stoix, stable-baselines3}. Additionally, while Jaxnasium algorithms can vectorize their training runs and shard them over multiple accelerators through JAX, its main focus remains simple setups for research-sized labs; large distributed (production-ready) reinforcement learning experiments are likely better solved with different libraries~\citep{liang2018rllib, toledo2024stoix, dekock2023mava}.

While our multi-agent transformation allows single-agent implementations to automatically operate on multi-agent environments, this is limited to individual multi-agent learning~\citep{dewitt2020ippo}. While we provide examples in our documentation for CTDE learning, complex multi-agent structures would require tweaking of the algorithms~\citep{rashid2020weightedqmixexpandingmonotonic}. 

Jaxnasium will be used as a research library. Future research will look into what engineering tricks have worked on various algorithms and how we can best operate over a wide variety of different tasks~\citep{lee2025simba, lee2025simbav2, hessel2018popart}. The aim is not to keep algorithms static, but rather implement newly found improvements to provide up-to-date algorithms, with proper defaults, that can operate on a wide variety of environments. The goal is that this provides researchers from various fields with access to a set of proper baselines and starting points for their research.

\section{Conclusion}

We presented Jaxnasium, a library aiming to make JAX-based reinforcement
learning more accessible and to broaden its use to environments not written in JAX, without code changes. Its central idea is to treat observation and action spaces as PyTrees
and to let the components around an algorithm, rather than the algorithm
itself, follow this PyTree structure. As a result, near-single-file implementations of various algorithms train on discrete, continuous, image and composite spaces, and even freely upgrade to the multi-agent setting, without changes to their code. These algorithm implementation files
serve as readable references for researchers as well as importable algorithms for practitioners. Through wrappers, they apply to over 300 registered environments, and additionally on host-based Gymnasium and EnvPool environments without sacrificing performance compared to algorithms designed for this setting.
Together with tools for scaffolding projects, multi-seed evaluation and
hyperparameter sweeps, we hope Jaxnasium provides researchers from various fields with a good starting point for running reinforcement learning agents on any type of environment.

\subsubsection*{Acknowledgments}
This work was financially supported by Shell Information Technology International Limited and the Netherlands Enterprise Agency under the grant PPS23-3-03529461.
We would lastly like to thank Jan Felix Kleuker for being a mathematician.

\bibliographystyle{iclr2027_conference}
\bibliography{biblio_non_anon}

\newpage

\appendix

\section{Included environments} \label{app:environment-list}

\begin{table}[h]
\centering
\small
\caption{Registered environments per suite currently available in Jaxnasium. For convenience, Jaxnasium bundles the 5 classic-control environments; other libraries have to be installed and can subsequently be initialized with \texttt{jym.make(...)}. More environments will likely be added as they become available.}
\label{tab:suites}
\begin{tabular}{llr}
\toprule
Suite & Domain & Envs \\
\midrule
Brax~\citep{brax2021github} & Continuous control & 10 \\
Craftax~\citep{matthews2024craftax} & Open-ended survival & 4 \\
Gymnax~\citep{gymnax2022github} & Classic control, bsuite, MinAtar & 24 \\
JaxMARL~\citep{flair2024jaxmarl} & Multi-agent coordination & 29 \\
Jumanji~\citep{bonnet2024jumanji} & Combinatorial problems & 26 \\
Navix~\citep{pignatelli2024navix} & Grid worlds & 41 \\
Octax~\citep{radji2025octax} & Arcade games (CHIP-8) & 40 \\
Pgx~\citep{koyamada2023pgx} & Board games, MinAtar & 20 \\
Playground~\citep{mujoco_playground_2025} & Robotics & 50 \\
PopJym~\citep{morad2023popgym} & Partial observability & 40 \\
XLand-MiniGrid~\citep{nikulin2023xlandminigrid} & Meta-RL grid worlds & 44 \\
Jaxnasium & Classic control & 5 \\
\midrule
Total & & 333 \\
\bottomrule
\end{tabular}
\end{table}

\section{Example of a sweep configuration} \label{app:sweep}

Sweeps are composed from search stages: grid search, one-at-a-time search,
random search and quasi-random Sobol search. Stages can be nested, for instance
a grid over environments with a random search over learning rates inside it, and
can branch per algorithm, so that each algorithm is searched over its own
hyperparameters (Listing~\ref{lst:sweep}). A sweep is a list of independent jobs that can be run in a loop; or, more typically, indexed by a Slurm job array on a cluster.

\begin{listing}[h]
\begin{minted}[xleftmargin=1em,numbersep=6pt]{python}
from jaxnasium.eval import AlgorithmEvaluation, GridSearch, SobolSearch, Sweep

evaluation = AlgorithmEvaluation(jax.random.split(key, 10))  # 10 seeds
sweep = Sweep(
    evaluation,
    GridSearch({"env": ["CartPole-v1", "Acrobot-v1"]}),
    GridSearch({"algorithm": {
        "PPO": [{"algorithm": PPO()},
                SobolSearch({"clip_coef": (0.1, 0.3)}, 32, seed=key)],
        "SAC": [{"algorithm": SAC()},
                SobolSearch({"tau": (1e-3, 5e-2)}, 32, seed=key)],
    }}),
)
result = sweep[job_index]()  # e.g. job_index = $SLURM_ARRAY_TASK_ID
\end{minted}
\caption{A sweep over two environments and two algorithms, each with its own
Sobol search, evaluated over ten seeds per configuration. 
\texttt{AlgorithmEvaluation} is a pre-implemented function for running a standard train+evaluate loop. However, any function can be defined and swept over.}
\label{lst:sweep}

\end{listing}

\newpage

\section{Extra environments for backend performance comparison} \label{app:performance-extra}

The figures below provide extra benchmarking runs over various backends, comparable to Figure~\ref{fig:benchmark-gymnasiumwrapper}. Figure~\ref{fig:benchmark-gymnasiumwrapper} was an average over only 6 simple environments (the 5 classic-control environments and Catch), whereas the environments below are more complex but do not have an implementation ready in each of the three backends (JAX-native, Gymnasium, EnvPool). Overall, these results demonstrate a similar picture: running the JAX-native version is typically fastest, but when the environment is not available in JAX, using our wrapper does not lead to diminished performance compared to using a JAX agent that is not built for end-to-end training in JAX.

\begin{figure}[h]
  \centering
  \includegraphics[width=0.8\linewidth]{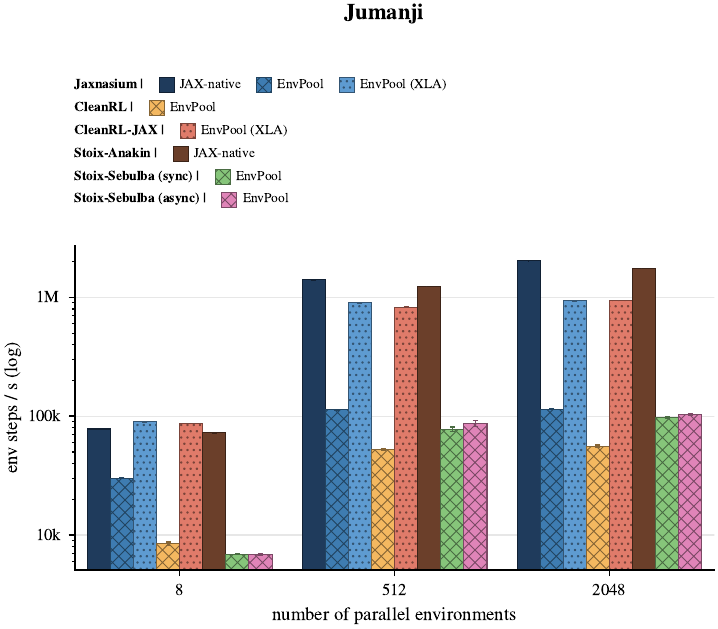}
  \caption{Throughput tested for various backends and different libraries on the Jumanji suite. Averaged over 5 runs. Standard deviation is plotted, but is negligible.}
\end{figure}

\begin{figure}[h]
  \centering
  \includegraphics[width=0.8\linewidth]{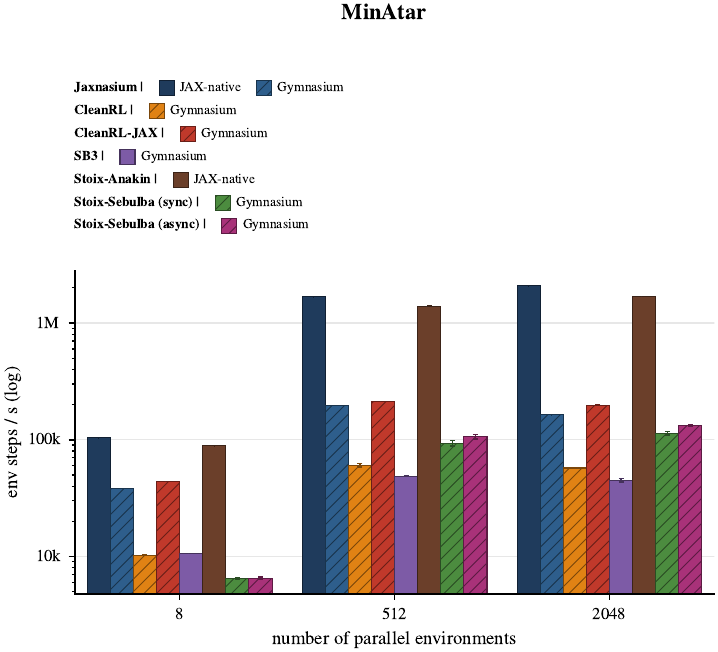}
  \caption{Throughput tested for various backends and different libraries on the MinAtar suite. Averaged over 5 runs. Standard deviation is plotted, but is negligible.}
\end{figure}

\begin{figure}[h]
  \centering
  \includegraphics[width=0.8\linewidth]{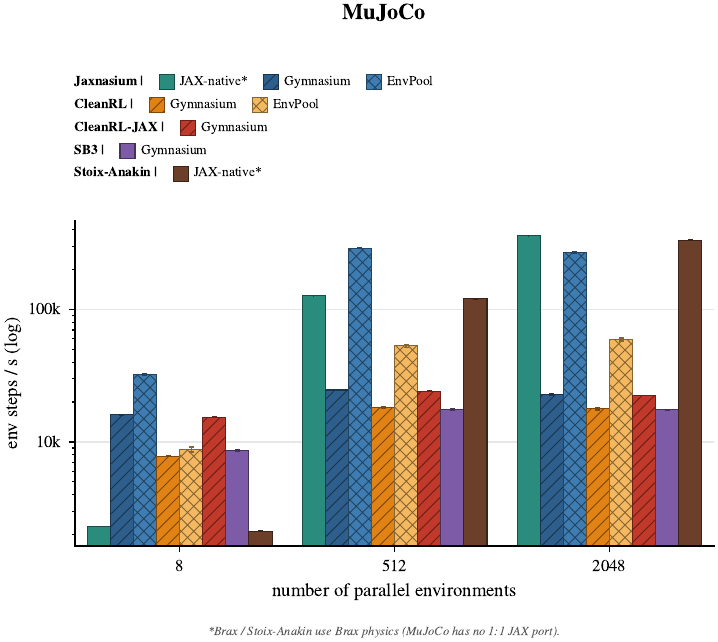}
  \caption{Throughput tested for various backends and different libraries on the MuJoCo and Brax suites. Averaged over 5 runs. Standard deviation is plotted, but is negligible.}
\end{figure}

\end{document}